\documentclass{article}
\usepackage{ijcai21}

\usepackage{times}
\usepackage{soul}
\usepackage{url}
\usepackage[hidelinks]{hyperref}
\usepackage[utf8]{inputenc}
\usepackage[small]{caption}
\usepackage{graphicx}
\usepackage{amsmath}
\usepackage{amsthm}
\usepackage{booktabs}
\usepackage{algorithm}
\usepackage{algorithmic}
\usepackage{enumerate}
\usepackage{amsfonts}
\usepackage{multirow}
\usepackage{multicol}
\usepackage{makecell}
\usepackage{float}
\usepackage{lipsum}
\title{Weakly Supervised Dense Video Captioning via Jointly Usage of Knowledge Distillation and Cross-modal Matching}

\author{
Bofeng Wu$^{1,2}$
\and
Guocheng Niu$^2$\and
Jun Yu$^{1}$\footnote{Jun Yu is the corresponding author}\and
Xinyan Xiao$^2$\and
Jian Zhang$^3$\And
Hua Wu$^2$
\affiliations
$^1$School of Computer Science and Technology, Hangzhou Dianzi University, Hangzhou, China\\
$^2$Baidu Inc., Beijing, China\\
$^3$Zhejiang International Studies University, Hangzhou, China
\emails
\{wubofeng, yujun\}@hdu.edu.cn,
\{niuguocheng, xiaoxinyan,wu\_hua\}@baidu.com,
jayzhang@outlook.com
}

\begin{document}

\maketitle

\begin{abstract}
This paper proposes an approach to Dense Video Captioning (DVC) without pairwise event-sentence annotation. First, we adopt the knowledge distilled from relevant and well solved tasks to generate high-quality event proposals. Then we incorporate contrastive loss and cycle-consistency loss typically applied to cross-modal retrieval tasks to build semantic matching between the proposals and sentences, which are eventually used to train the caption generation module. In addition, the parameters of matching module are initialized via pretraining based on annotated images to improve the matching performance. Extensive experiments on ActivityNet-Caption dataset reveal the significance of distillation-based event proposal generation and cross-modal retrieval-based semantic matching to weakly supervised DVC, and demonstrate the superiority of our method to existing state-of-the-art methods.
\end{abstract}
\section{Introduction}
Dense Video Captioning (DVC)~\cite{krishna2017dense} refers to detecting and describing multiple events from a given video. Each event is represented by a video segmentation reflecting distinct semantic with starting and ending timestamps in the video and the semantic should be expressed by a sentence constructed by lexical words. Being able to deliver fine-grained semantic of a long video sequence, DVC boosts the applications of techniques like content-based video retrieval, personalized recommendation and intelligent video surveillance. Most of the state-of-the-art approaches to DVC adopt a fully supervised learning pipeline~\cite{krishna2017dense,zhou2018end,xiong2018move,mun2019streamlined}, which means in addition to the videos, the lexical description of each involved event with its starting and ending timestamps should also be known before training. However, locating events in each video and tagging each event with appropriate sentence is extremely labor-intensive, given mass amount of videos.

In the past several years, weakly supervised DVC methods~\cite{shen2017weakly,duan2018weakly} emerged to achieve event localization and sentence captioning for each event, which only needs videos as well as lexical descriptions of the whole video sequence as training data. Therefore, weakly supervised DVC is much more challenging than supervised DVC due to the lack of event-level annotation. Shen et al.~\cite{shen2017weakly} mapped sentence lexical words to frame regions and constructed region-based sequences, then associated lexical words to sequences. Region-based sequences describe more object-level than event-level semantic, so this method is insufficient to generate proposals of events. Duan et al.~\cite{duan2018weakly} firstly extracted video and caption features using RNNs, then associated these features by cross-attention multi-modal fusion process for event clip generation, the training of captioning module simultaneously exploits event clips and captions. Their method needs captions to localize clips, but the captions are unavailable during the testing phase, so their clips only rely on random generation, which is inaccurate. So far as we know, there is still no satisfactory solution to event-caption generation in current weakly supervised DVC.

This paper proposes a novel weakly supervised DVC pipeline towards high-quality event proposal generation and accurate caption generation. This pipeline consists of three modules: the distillation-based proposal generation, the proposal-caption matching by cross-modal retrieval and the event caption generation. The proposal generation module detects a group of candidate proposals for event depiction, and is optimized with soft label constraints and intermediate feature constraints, which are distilled from several teacher networks designed for similar tasks. The proposal-caption matching module selects one proposal for each sentence to formulate pairwise data. The matching module individually extracts video and sentence features through the encoder of a transformer followed by an attention block, and performs semantic alignment between visual and lexical features using contrastive loss and cycle-consistency loss. Apart from this, images with captions are used to construct pseudo dense video with captions to pretrain the matching module to obtain good initialization parameters. The event caption generation module adopts Attention Long Short-Term Memory (Attention-LSTM)~\cite{anderson2018bottom} to achieve the sentence generation for selected proposals. Compared with end-to-end methods, each module in our pipeline is trained independently, which enables performance promotion of the pipeline via individual optimization of each module. In the meanwhile, positive interactions between different modules also improve the performance of the pipeline. For example, high-quality proposal-sentence matching improves the proposal and caption generation results, while good proposal generation benefits the matching module.

The contributions of this paper are three-fold:
\begin{itemize}
\item We propose an efficient pipeline to tackle the weakly supervised DVC task. In particular, our pipeline innovatively introduces a cross-modal matching module to achieve proposal-caption matching, which is currently ignored by other methods.  With the help of the proposed matching module, the remaining two modules (i.e. the proposal generation module and the caption generation module) are effectively integrated to better deal with this task. We conduct comprehensive experiments to empirically analyze our method on Activity-Caption dataset. The experimental results demonstrate the superiority of our proposed approach to existing state-of-the-art methods.
\item We adopt knowledge distillation techniques to conduct event proposal generation. By distilling the knowledge from other well trained teacher networks and using the well designed learning strategy, the proposal generation module can make full use of proposal data from other fields and it is quiet flexible. Experiments show that our multi-teachers knowledge distillation learning method achieves outstanding results.
\item We introduce a novel cross-modal retrieval mechanism into the proposal-caption matching. Specifically, we exploit contrastive loss and cycle-consistency loss to optimize cross-modal matching. It is worthy noting that we also improve matching performance via pretraining the network using preprocessed images with ground truth captions.
\end{itemize}
\section{Related Works}
\subsection{Proposal Generation}
The event proposal generation plays an important role in the DVC tasks. ~\cite{krishna2017dense} adopted Deep Action Proposals (DAP) model~\cite{escorcia2016daps}, which generated variable-length proposals based on the clustered ground truth annotation. ~\cite{xiong2018move} exploited Structured Segment Network (SSN)~\cite{zhao2017temporal}, which boosted the accuracy of proposals by modeling their temporal structure via a structured temporal pyramid. Similarly, ~\cite{zhou2018end} used multi-layer transformer encoder to generate proposals from visual features. Single-Stream Temporal (SST)~\cite{buch2017sst} adopted in ~\cite{mun2019streamlined} constructed candidate proposals using GRU hidden states with various lengths and individual confidence score for proposal refinement. \emph{However, these methods usually cannot generate proposals with precise starting and ending boundaries.} Bound Sensitive Network (BSN) proposed in ~\cite{lin2019bmn} defined candidate proposal boundaries as frames with high probability and adopted confidence score to achieve proposal generation. The state-of-the-art Boundary-Matching Network (BMN)~\cite{lin2019bmn} further developed BSN into an end-to-end framework.

In weakly supervised DVC scenario, the lacking of event timestamps annotation turns proposal generation into an unsupervised problem, therefore the above mentioned methods are invalid under this challenging situation. Integrating the knowledge supplied from other similar tasks can mitigate this problem, which can be regarded as a knowledge distillation strategy~\cite{hinton2015distilling}.
\subsection{Proposal Matching}
Given generated proposals as well as sentence-level captions obtained from annotation, the proposal caption generation is still unavailable because there is no correspondence between the proposals and captions. To this end, a cross-modal matching process is necessary, which often finds its usage in cross-modal retrieval task.

~\cite{peng2019comic} proposed to project the data of different modalities into one common space, in which the projection embraces the geometric consistency (GC) and the cluster assignment consistency (CAC). ~\cite{song2019polysemous} extracted modality-specific features through self-attention and residual learning. The model was optimized by minimizing the distance between relevant video and text representations. ~\cite{gabeur2020multi} adopted multi-modal transformer and BERT to learn video and text features respectively, and used bi-directional loss to rank the cross-modal similarity. ~\cite{ging2020coot} employed transformer and attention for modality-specific feature extraction, and performed multi-level cross-modal semantic alignment reinforced by a cross-modal cycle-consistency loss, which enabled this approach to yield state-of-the-art results. ~\cite{yu2020long} proposed a memory attention mechanism to identify the critical visual representations related to the language representations, which also similar to multi-modal matching.
\section{The Proposed Method}
\begin{figure*}
  \centering
  \includegraphics[totalheight=2.7 in]{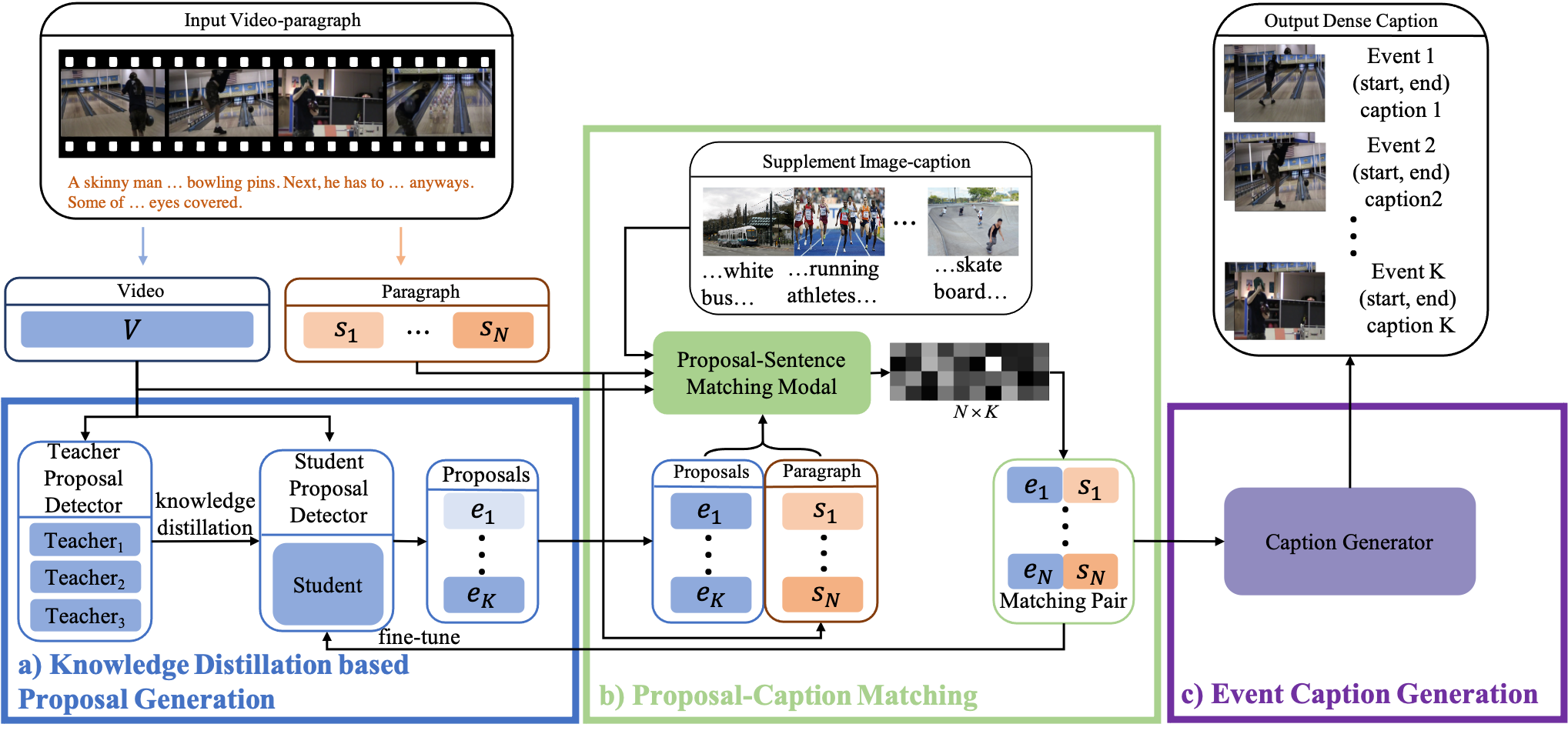}
  \caption{Overview of the proposed pipeline: a) Proposal Generation detects valid event proposals based on Knowledge Distillation in an untrimmed video; b) Proposal-caption Matching selects one event proposal for each sentence by matching between all event proposals and sentences; c) Event Caption Generation module generates captions for each proposal.} \label{fig:graph1}
\end{figure*}

Weakly supervised DVC detects $N$ video clips $\{c_i\}_{i=1}^N$ (each can be described as a proposal with both visual features and starting/ending timestamps) corresponding to certain events from a whole video sequence $V$, and describes $c_i$ with a lexical sentence $s_i=\{w_i^j\}_{j=1}^{M_i}$ containing $M_i$ words $w_j$. The training data in weakly supervised DVC task only include full video sequences and their lexical descriptions called paragraphs. In the rest part of the paper, we use $V$ and $c$ to represent visual features ($V$ denotes video-level features and $c$ denotes clip-level features), and adopt $s$ to describe lexical features. The proposed method is presented in Figure \ref{fig:graph1} which consists of three modules: a) knowledge distillation based proposal generation (KDPG, Section \ref{kdpg}) generates candidate event proposals about certain events in the video; b) proposal-caption matching(PCM, Section \ref{psm}) selects a corresponding event proposal from candidates for each sentence; and c) event caption generation (ECG, Section \ref{ecg}) generates caption for each event proposal.

\subsection{Knowledge Distillation based Proposal Generation}
\label{kdpg}
Generating accurate event proposal is very difficult in weakly supervised scenario. This enlightens us to introduce knowledge learned from relevant problems that have been well solved. Specifically, we resort to network distillation. We train several teacher networks on other proposal detection tasks (in this paper, i.e. activity, action, and video highlight ) in fully supervised way, and use soft labels as well as intermediate features produced by teacher networks to train our student network.
\subsubsection{Teacher Network} modifies the BMN model proposed in ~\cite{lin2019bmn} via changing the CNN into Temporal Convolutional Network (TCN)~\cite{lea2017temporal} to improve the temporal information extraction ability. We train teacher networks on the aforementioned three tasks. Given a video, teacher networks output soft-labels $\{O_i\}_{i=1}^n$ ($n$ is the teacher number) comprising probability scores that each frame being a starting/ending point of a proposal, and the confidence scores that each pair of the potential starting and ending frames being a proposal. These proposals are selected according to the ranking of their scores computed based on probability scores and confidence scores. Also, the teacher networks extracts visual encoding features $\{V_i\}_{i=1}^n$.

\subsubsection{Student Network} adopts the same network structure as teachers. In training, student network learns visual features $V'$ and labels $O'$ under the guidance of soft labels $\{O_i\}_{i=1}^n$ and intermediate features $\{V_i\}_{i=1}^n$ produced by teachers. We compute a weight $\{g_i\}_{i=1}^n$ for each teacher to leverage its contribution to the student:
\begin{equation}
\begin{cases}
g_i=\left. e^{q_i} \middle/ \sum_{j=1}^ne^{q_j} \right.\\
q_i = \mbox{FC}(\sigma(\mbox{MaxPool}(\sigma(V_i\cdot V'))))
\end{cases},
\end{equation}
where FC represents fully connected layer and $\sigma(x)=\mbox{ReLU}(\mbox{FC}(x))$. Then, we train the student network via $L_p$
\begin{equation}
L_p=f(O',\sum^n_{i=1} g_i O_i) + \sum^n_i(g_i\cdot\mbox{MSE}(V',V_i)),
\end{equation}
where $f$ is loss function used in ~\cite{lin2019bmn} and MSE is mean squared error loss function, we detail $f$ in supplementary material. The student network follows the same proposal score computation method adopted by teacher networks.

In order to prevent the student network from generating a too high weight for one teacher network, we add suppression weights $\{\bar{g}_i=\frac{1}{n\cdot g_i}\}_{i=1}^n$ inspired by weighted binary cross entropy loss function to compute suppression loss $\bar{L}_p$:
\begin{equation}
\bar{L}_p=f(O',\sum^n_{i=1} \bar{g}_i O_i) + \sum^n_i(\bar{g}_i\cdot\mbox{MSE}(V',V_i)),
\end{equation}
and the final loss function of the event proposal generation is calculated as:
\begin{equation}
L_p^{final} = \gamma\cdot L_p+\eta\cdot \bar{L}_p,
\end{equation}
where $\gamma$ and $\eta$ are hyperparameters and we set as 0.8, 0.2 after several experiments on the premise of avoiding the excessive role of the suppression loss.

We compute scores of all candidate proposals that exist in a video, and select top $K$ proposals as valid event proposals $E=\{e_i\}_{i=1}^K$, the proposal score computation is detailed in supplementary material.
\subsection{Proposal-Caption Matching}
\label{psm}
We adopt COOT~\cite{ging2020coot} for weakly supervised proposal-caption matching. In original model, given ground truth event-level annotation, COOT uses contrastive loss $L_{con}$ to enforce the relevant (positive) visual-lexical pair are close while irrelevant (negative) pair are far from each other. Through COOT, visual modal generates three-level(video, event and contextual) representation, and lexical modal also generates three-level(paragraph, sentence and contextual) representation corresponding to visual modal. Our training data only contains video-level annotation so we can not construct event and contextual level representation in visual modal. Owing to the absent of event-level annotations, we evenly divide a video into multiple clips such that each clip corresponds to a sentence in the paragraph. Thus, given a sentence, the corresponding clip can be the sentence’s positive clip and the rest are the negative clips. We create the positive and negative sentences in the same way.

Also, to enhance the matching performance, COOT exploits cycle-consistency loss $L_{cycle}$ to guarantee the matched event to a sentence can be used to locate the same sentence, and vice versa. Similarly, we use evenly divided events to substitute the true event for training. Note that matching aims to generate sentence-proposal(i.e. pseudo event-level annotation) pairwise data that will be used to train the subsequent caption generation module and refine the proposal generation module (detailed in supplementary material). In the testing phase that we showed in Fig \ref{fig:graph2}, we can neglect the matching module and feed the generated event proposals to caption generation module directly because the two generation modules have been already well-trained on pairwise data. Below are the two losses adopted in our training:
\begin{figure*}
  \centering
  \includegraphics[totalheight=1.4in]{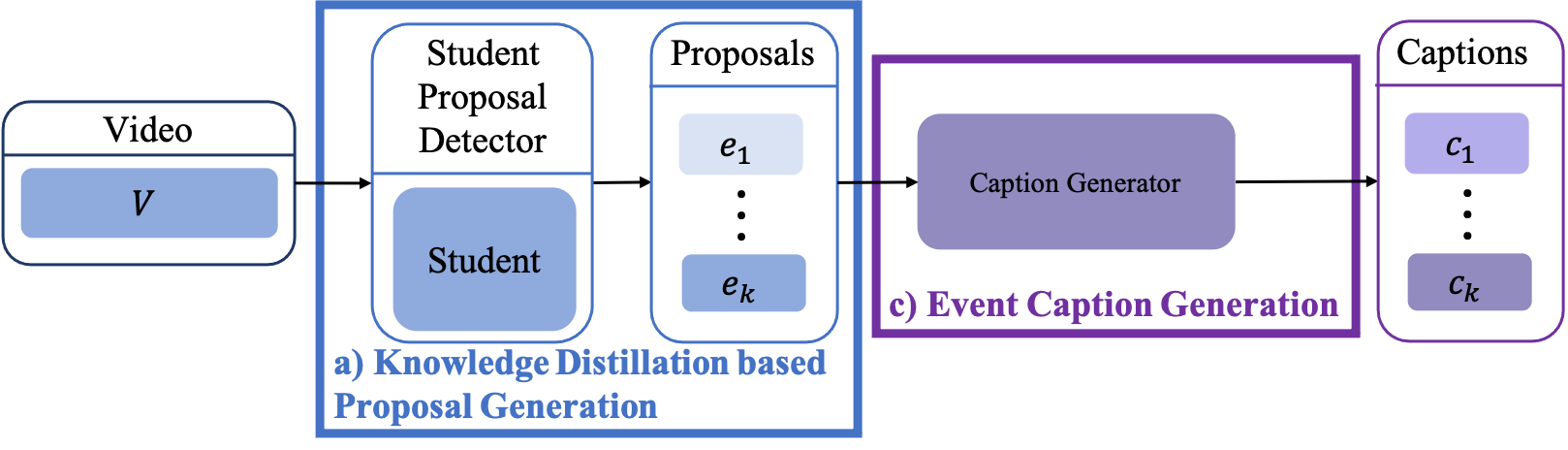}
  \caption{In testing phase we utilize Knowledge Distillation based Proposal Generation and Event Caption Generation to generate dense captions. } \label{fig:graph2}
\end{figure*}
\subsubsection{Contrastive Loss}
COOT adopts contrastive loss on all three-level visual-lexical(video-paragraph, event-sentence, contextual-contextual) representation and losses share the same form, we only explain the event-sentence level loss. Given a positive pair $(c^+,s^+)$ and two negative pairs ${(c^-,s^+),(c^+,s^-)}$, $L_{con}$ is defined as:
\begin{align}
\begin{split}
L_{con} &= \sum L((c^+,s^+),\{(c^-,s^+),(c^+,s^-)\},h) \\
        &+ \sum L((1,1),\{(c^+,c^-),(s^+,s^-)\},h)
\end{split},
\end{align}
where $h$ is a margin hyperparameter, the $(1,1)$ pair denotes that positive samples are not changed, and
\begin{align}
\begin{split}
L((c^+&,s^+),\{(c^-,s^+),(c^+,s^-)\},h) \\
      &=\mbox{max}(0,h+D(c^+,s^+)-D(c^-,s^+)) \\
      &+\mbox{max}(0,h+D(c^+,s^+)-D(c^+,s^-))
\end{split},
\end{align}
where $\left. D(c^+,s^+)=1-({c^+})^Ts^+ \middle/ (||c^+||||s^+||) \right.$ is the cosine distance between two vectors.

\subsubsection{Cycle Consistency Loss}
Given a sentence $s_i$, we first compute its clip counterpart $\bar{c}_{s_i}$, and then cycle back to the sentence sequence $\{s_i\}_{i=1}^N$ and calculate the soft location $u$:
\begin{equation}
\begin{cases}
\bar{c}_{s_i}=\sum^N_{j=1} \alpha_j c_j &\mbox{where } \alpha_j=\frac{e^{-||s_i-c_j||^2}}{\sum^N_{k=1}e^{-||s_i-c_k||^2}} \\
u=\sum^N_{j=1} \beta_j j &\mbox{where } \beta_j=\frac{e^{-||\bar{c}_{s_i}-s_j||^2}}{\sum^N_{k=1}e^{-||\bar{c}_{s_i}-s_k||^2}}
\end{cases},
\end{equation}
$\alpha_j$ is the similarity of clip $c_j$ to sentence $s_i$, and $\beta_j$ is the similarity of $s_j$ to $\bar{c}_{s_i}$. The object of $L_{cycle}^{sent}$ is to reduce the distance between the source location $i$ and the soft location $u$:
\begin{equation}
L_{cycle}^{sent} = ||i-u||^2.
\end{equation}
Similarly, we conduct the cycle consistency evaluation given a clip representation $c_i$ by $L_{cycle}^{clip}$. Then, the cycle-consistency loss is defined as $L_{cycle}=L_{cycle}^{sent}+L_{cycle}^{clip}$.

The final matching loss $L_m$ is defined as follows:
\begin{equation}
L_m=L_{con}+L_{cycle}.
\end{equation}

\subsubsection{Pretraining based on Annotated Images}
We further propose to enhance the matching module using images with ground truth captions. We use annotated images because image is the basic unit of video and the annotation is widely available. We believe annotated images can provide additional static information for building the match between video clips and sentences.

Specifically, given images $\{x_i\}_{i=1}^n$ with captions $\{y_i\}_{i=1}^n$, we simulate pseudo video by duplicating these images and adding Gaussian noise. The pseudo paragraph is constructed by concatenating corresponding captions. We use pseudo data to pretrain COOT to obtain good initialization parameters before training with true videos and paragraphs.

\subsection{Event Caption Generation}
\label{ecg}
The contribution of this paper is on how to use knowledge distillation and cross-modal matching to design a pipeline for weakly supervised DVC. In order to highlight the advantages of our pipeline rather than the advanced sub-model, we use the widely used Attention-LSTM network, then train it on the generated pairwise data to obtain the sentence description for each generated proposal. Attention-LSTM includes two LSTM layers and one attention layer to encode input proposal features and decode to a sentence. At each time step, Attention-LSTM uses previous hidden state and generated word to generate a word probability vector, which we detail in supplementary material. We apply cross-entropy loss $L_c$ as follows to minimize the distance between the one-hot vector of ground-truth caption $s=\{w_i\}_{i=1}^M$ and our prediction $\bar{s}=\{\bar{w}_i\}_{i=1}^M$:
\begin{equation}
L_c = -\sum^M_{t=1} w_t \cdot log(\bar{w}_t|w_1:w_{t-1}).
\end{equation}

\begin{table}[tb]
\centering
\begin{tabular}{ccc}
\hline
Method & AR@100(\%) & AUC(\%) \\
\hline
\hline
Self-Attn~\cite{zhou2018end} & 52.95 & - \\
\hline
$\rm TN_{activity}$ & 67.87 & 68.13 \\
$\rm TN_{action}$ & 63.65 & 68.18 \\
$\rm TN_{highlight}$ & 53.26 & 59.99 \\
\textbf{KDPG} & \textbf{69.38} & \textbf{69.86} \\
\hline
\end{tabular}
\caption{Comparison of event proposal generation performance between proposed KDPG and other methods on ActivityNet-Caption validation set.}
\label{tab:proposal table}
\end{table}

\begin{table*}
\centering
\begin{tabular}{ccccccc}
\hline
Method & M & C & B@1 & B@2 & B@3 & B@4 \\
\hline
\hline
DCEV(fully supervised) & 4.82  & 17.29 & 17.95 & 7.69 & 3.86 & 2.20 \\
SDVC(fully supervised) & 8.82  & 30.68 & 17.92 & 7.99 & 2.94 & 0.93 \\
\hline
WSDECV(weakly supervised) & 6.30 & 18.77 & 12.41 & 5.50 & 2.62 & 1.27 \\
\textbf{ECG(weakly supervised)} & 7.06 & 14.25 & 11.85 & 5.64 & 2.71 & 1.33 \\
\hline
\end{tabular}
\caption{Caption generation performance comparison of different full supervised and weakly supervised DVC method with our pipeline.}
\label{tab:caption table}
\end{table*}

\section{Experiments}
To demonstrate the effectiveness of the proposed pipeline, we conduct experiments on the dataset ActivityNet-Caption~\cite{krishna2017dense}. For knowledge-distillation, we adopt multiple external datasets used in the task of Temporal Proposal Detection (TPD), i.e, THUMOS-14~\cite{THUMOS14}, ActivityNet-Action~\cite{krishna2017dense}, BROAD-Video Highlights\footnote{\url{http://ai.baidu.com/broad/}}. For pretraining COOT, we utilize a typical Image-Captioning dataset MS-COCO~\cite{lin2014microsoft}, then we use ActivityNet-Caption to continue training COOT. Supplementary material summarizes these datasets.

We compare our method with several representative methods, such as classical DVC model proposed in ~\cite{krishna2017dense}, streamlined DVC (SDVC) method~\cite{mun2019streamlined} and weakly supervised DVC (WSDVC) method~\cite{duan2018weakly}. We also compare the event proposal generation module with Self-Attn ~\cite{zhou2018end} to demonstrate the effectiveness of knowledge distillation strategy.

\subsection{Datasets and Processing}
ActivityNet-Caption dataset has 20k untrimmed videos devided into training, validation and testing subsets by ratio 2:1:1, and each video on average contains 3.6 event clips with captions.
THUMOS-14 dataset has 0.4k untrimmed videos containing multiple action clips with timestamps. We split it into training and validation subsets by ratio 4:1.
ActivityNet-Action dataset shares same untrimmed videos as ActivityNet-Caption but has another type of labeled information for temporal action detection task.
BROAD-Video Highlights dataset has 1.4k untrimmed entertainment long video with average 12.2 highlights with timestamps. We split it into training and validation subsets by ratio 4:1.
MS-COCO dataset aims to solve image-captioning task, and we sampled 50k images with randomly selected one corresponding caption as our supplement fine-grained dataset.

In KDPG module, we extract video features using pretrained C3D~\cite{tran2015learning} network on each TPD dataset and reduce the dimensionality of output from 4096 to 500 using PCA. In PCM module, to guarantee the consistency of video and image features, we adopt ResNet to scan each frame in video and each image into a 2048-D vector on ActivityNet-Caption and MS-COCO. In ECG module, we extract video and lexical features on ActivityNet-Caption using video feature extraction method in ~\cite{ging2020coot} and pretrained BERT~\cite{devlin2018bert}.

\subsection{Comparison Baseline and Setup}
For event proposal generation, we follow ~\cite{lin2019bmn} to rescale the length of each feature sequence to 100 using linear interpolation. In the training of teacher networks, we use SGD optimizer and set learning rate to 0.001, batch size to 16. In the training of student network, we set learning rate, batch size, $\gamma$ and $\eta$ to 0.0005, 32, 0.8 and 0.2 respectively. In the caption generation of testing phase we use beam search strategy and set the beam size to 5. We mainly compare with Self-Attention (Self-Attn)~\cite{zhou2018end}, which uses video encoder with proposal decoder to generate event proposal.

For event caption generation, we use Adam optimizer and set learning rate to 0.02 and decays every 32 epochs with decay rate 0.1, batch size to 16. We compare our method with the following two fully supervised methods and one weakly supervised methods:
\begin{itemize}
\item[1)] Dense-Caption Events in Video (DCEV) proposed in ~\cite{krishna2017dense} used DAPs to generate event proposals and context-based caption generator to generate captions.
\item[2)] Streamlined Dense Video Captioning (SDVC) used SST~\cite{buch2017sst}+ESGN to generate event proposals and used a hierarchical episode-event RNN to generate caption.
\item[3)] Weakly Supervised Dense Event Captioning in Video (WSDECV)~\cite{duan2018weakly} used pretrained caption generator and sentence localizer to accomplish end-to-end generation.
\end{itemize}

\begin{table}[b]
\centering
\begin{tabular}{ccc}
\hline
Method & AR@100(\%) & AUC(\%) \\
\hline
\hline
Fully supervised BMN & 70.43 & 71.42 \\
Averaged weights & 67.89 & 68.13 \\
KDPG & 69.38 & 69.86 \\
\textbf{KDPG + pairwise} & \textbf{70.77} & \textbf{72.05} \\
\hline
\end{tabular}
\caption{Ablation results of event proposal generation module when using different strategies.}
\label{tab:proposal ablation table}
\end{table}

\begin{table*}
\centering
\begin{tabular}{ccccccc}
\hline
Method & M & C & B@1 & B@2 & B@3 & B@4 \\
\hline
\hline
Vanilla & 6.51  & 12.83 & 11.31 & 5.31 & 2.49 & 1.20 \\
ECG w/o & 6.88 & 13.92 & 11.71 & 5.52 & 2.68 & \textbf{1.39} \\
\textbf{ECG} & \textbf{7.06} & \textbf{14.25} & \textbf{11.85} & \textbf{5.64} & \textbf{2.71} & 1.33 \\
\hline
\end{tabular}
\caption{Comparison of event captioning results with different training corpus.}
\label{tab:caption ablation table}
\end{table*}

\begin{table}[htb]
\centering
\begin{tabular}{ccc}
\hline
Method & R@1(\%) & MR \\
\hline
\hline
PCM w/o      & 52.6 & 1.8 \\
\textbf{PCM} & \textbf{53.8} & 1.8 \\
\hline
\end{tabular}
\caption{The matching performance comparison of whether using image-captioning data to pretrain the PCM module.}
\label{tab:matching ablation table}
\end{table}

\subsection{Experimental Results}
Table \ref{tab:proposal table} shows the proposal generation performance comparison of our KDPG module with other methods and every single teacher network (TN) we used. To evaluate the quality of a given proposal, we use the proposal to retrieve true event clips and compute the Average Recall (AR) under multiple IoU thresholds [0.5:0.05:0.95]. Then we calculate the mean AR about the top 100 generated proposals (AR@100). In addition, we calculate the Area under the AR vs. AN (Average Number of proposals) curve (AUC). These results are calculated from the validation set in Activity-Caption dataset. The reported results in Table \ref{tab:proposal table} demonstrate the outstanding proposal generation capability of KDPG. Due to the superior global-context extraction ability of BMN+TCN and the appropriate distillation strategy we used, KDPG outperforms the Self-Attn and three teacher networks.

Table \ref{tab:caption table} shows the caption generation performance comparison of DVC metrics among the DCEV, SDVC, WSDECV and our ECG. To measure the performance of the captioning results, we use commonly-adopted evaluation metrics, i.e. METEOR(M), CIDEr(C) and BLEU@N. We generate event proposals based on different IoU thresholds of [0.3,0.5,0.7,0.9], and compute average metrics over the threshold using official codes\footnote{\url{https://github.com/ranjaykrishna/densevid_eval}}. The presented results in Table \ref{tab:caption table} illustrate that the performance of ECG has comprehensively exceeded the weakly supervised method WSDECV. In some metrics, ECG even exceeds fully supervised methods. We believe the reason is two-fold: First, the distillation strategy produces much better proposals than other methods; Second, our cross-modal matching module builds good correspondences between proposals and sentence captions, which is unavailable in other methods.

\begin{figure}[tb]
  \centering
  \includegraphics[totalheight=4.6in]{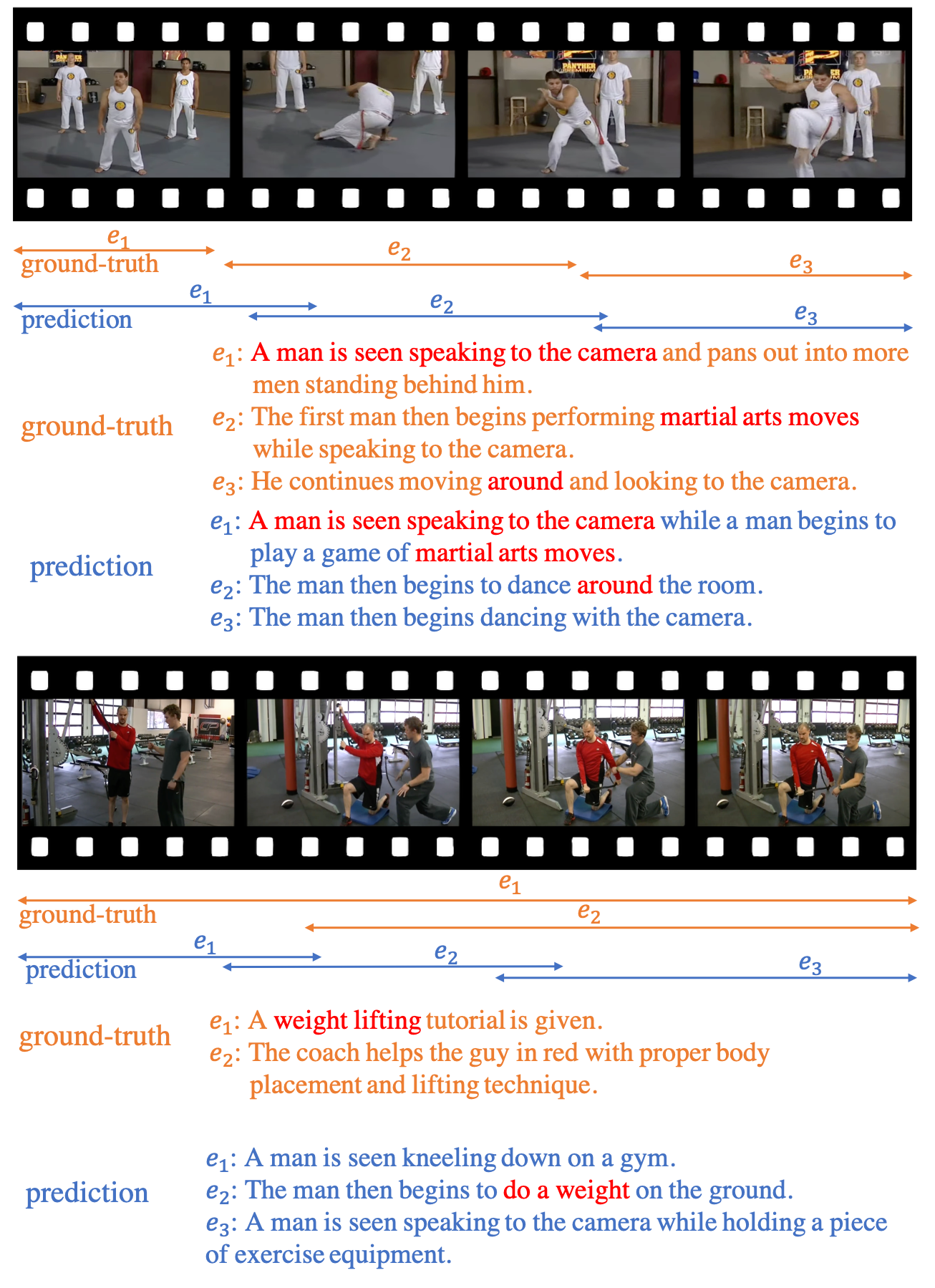}
  \caption{Qualitative results on ActivityNet Captions.} \label{fig:graph3}
\end{figure}

Figure \ref{fig:graph3} shows the qualitative results of our method KDPG-DVC. It is worth mentioning that different events in the same video may have different descriptions, such as "martial arts moves" and "dance", the reason is that each video segment input into the captioner is equivalent to a separate short video. The generated caption is also independent and the context is not strongly related. Although this issue is not our focus, we will consider an appropriate solution in our future work.

\subsection{Ablation Study}
In the ablation study, we alter the training pipeline of our model by overlying different training steps to justify the effects of different modules.

Table \ref{tab:proposal ablation table} demonstrates the effectiveness of our novel multi-teachers knowledge distillation learning for event proposal generation. In detail, fully supervised BMN method means BMN trained on fully supervised data. The averaged weights method stands for averaging the weights of all teacher networks. The method of KDPG plus pairwise data denotes that we simultaneously use event annotation of pairwise proposal-sentence data generated by PCM module as hard labels. Experimental results show that self-adaptive weight learning in KDPG is much better than the averaged weights method, and this is because adaptive weight computation transfers appropriate knowledge from all kinds of teachers to the student. In addition, we find the PCM module to be a great help to event proposal generation. With the supervision of pairwise data generated by PCM module, KDPG has been significantly improved and even outperforms the fully supervised method. This demonstrated that our pairwise data supplements more event information that cannot be completely learned from the event annotated data. With our knowledge distillation method, the video clip information of different types is greatly learned, which can help the event extraction. From experience, we believe that iterating multiple times this training strategt can improve the performance of the model, but this is not an innovation of our method, we only iterated once.

Besides, we conduct another experiment to confirm that our PCM module greatly benefits the final event caption generation as shown in Table \ref{tab:caption ablation table}. The method of vanilla indicates that the event caption generator is trained on the existing video-paragraph data corpus, which is the only data we can use without our PCM module. While the ECG w/o means the training data is automatically generated by PCM module. We believe the PCM module provides more precise and general visual-language semantic information via cross-modal matching, which indeed helps a lot. Our final ECG method takes full advantage of both. Specifically, we get event proposals by KDPG+pairwise model and generate pairwise annotated data by PCM module next. Then We train ECG with video-paragraph data firstly and refine it using pairwise annotated data.

To prove that pretraining with image-captioning data promotes the matching capability of PCM module, we report the commonly used retrieval metrics (i.e. R@1 and Median Rank (MR)) calculated on the validation set of ActivityNet-Caption in Table \ref{tab:matching ablation table}. Specifically, the method of PCM w/o is the basic version of our PCM module. It can only depend on limited number of coarse-grained training corpus, including the video-paragraph level and the constructed event-sentences level data. While in the approach of PCM, more accessible and fine-grained image-captioning data is utilized by the proposal-caption matching network for pretraining. Experiments show that the R@1 of PCM outperforms PCM w/o by 1.2\%, which demonstrates the effectiveness of the pretraining phase.

\section{Future Work}
In our future work, we consider conducting comparative analysis research about the strength and weaknesses of pipeline and end-to-end architectures to decide whether we convert our model architecture from pipeline to end-to-end.

Limited by the number of knowledge datasets, our knowledge distillation module has not reached the optimal state, we will continue to investigate related datasets to enhance the performance. And we believe use other caption generation models also can help our pipeline a lot, especially in enhancing the context relation between different events in the same video.

It is also important to note that the usage of labeled images is simple, and we will consider how to better integrate the information of static images as a supplement into the video encoding and decoding phase.

\section{Conclusion}
In this paper, we present an efficient pipeline which contains three modules i.e., distillation learning based proposal generation, proposal-caption matching and event caption generation addressing the weakly supervised DVC task. Knowledge distillation learning is used to solve the unsupervised proposal generation task and cross-modal matching is used to generate precise proposal-sentence pairs from video-paragraph. Joint usage of the above-mentioned methods solves the proposal generation and event-caption generation challenges of weakly supervised DVC. This pipeline architecture can provide promotions to the full pipeline by improving every single module, and the positive interaction between every module also promotes the full pipeline.

Experimental results on the dataset of ActivityNet-Caption demonstrate the significance of distillation-based event proposal generation and cross-modal retrieval-based semantic matching to weakly supervised DVC.

\section*{Acknowledgements}
This work was supported by National Natural Science Foundation of China under Grant 61836002.

\bibliographystyle{named}
\bibliography{ijcai21}

\begin{thebibliography}{}

\bibitem[\protect\citeauthoryear{Anderson \bgroup \em et al.\egroup
  }{2018}]{anderson2018bottom}
Peter Anderson, Xiaodong He, Chris Buehler, Damien Teney, Mark Johnson, Stephen
  Gould, and Lei Zhang.
\newblock Bottom-up and top-down attention for image captioning and visual
  question answering.
\newblock In {\em Proceedings of the IEEE conference on computer vision and
  pattern recognition}, pages 6077--6086, 2018.

\bibitem[\protect\citeauthoryear{Buch \bgroup \em et al.\egroup
  }{2017}]{buch2017sst}
Shyamal Buch, Victor Escorcia, Chuanqi Shen, Bernard Ghanem, and Juan
  Carlos~Niebles.
\newblock Sst: Single-stream temporal action proposals.
\newblock In {\em Proceedings of the IEEE conference on Computer Vision and
  Pattern Recognition}, pages 2911--2920, 2017.

\bibitem[\protect\citeauthoryear{Devlin \bgroup \em et al.\egroup
  }{2018}]{devlin2018bert}
Jacob Devlin, Ming-Wei Chang, Kenton Lee, and Kristina Toutanova.
\newblock Bert: Pre-training of deep bidirectional transformers for language
  understanding.
\newblock {\em arXiv preprint arXiv:1810.04805}, 2018.

\bibitem[\protect\citeauthoryear{Duan \bgroup \em et al.\egroup
  }{2018}]{duan2018weakly}
Xuguang Duan, Wenbing Huang, Chuang Gan, Jingdong Wang, Wenwu Zhu, and Junzhou
  Huang.
\newblock Weakly supervised dense event captioning in videos.
\newblock In {\em Advances in Neural Information Processing Systems}, pages
  3059--3069, 2018.

\bibitem[\protect\citeauthoryear{Escorcia \bgroup \em et al.\egroup
  }{2016}]{escorcia2016daps}
Victor Escorcia, Fabian~Caba Heilbron, Juan~Carlos Niebles, and Bernard Ghanem.
\newblock Daps: Deep action proposals for action understanding.
\newblock In {\em European Conference on Computer Vision}, pages 768--784.
  Springer, 2016.

\bibitem[\protect\citeauthoryear{Gabeur \bgroup \em et al.\egroup
  }{2020}]{gabeur2020multi}
Valentin Gabeur, Chen Sun, Karteek Alahari, and Cordelia Schmid.
\newblock Multi-modal transformer for video retrieval.
\newblock In {\em European Conference on Computer Vision (ECCV)}, volume~5.
  Springer, 2020.

\bibitem[\protect\citeauthoryear{Ging \bgroup \em et al.\egroup
  }{2020}]{ging2020coot}
Simon Ging, Mohammadreza Zolfaghari, Hamed Pirsiavash, and Thomas Brox.
\newblock Coot: Cooperative hierarchical transformer for video-text
  representation learning.
\newblock {\em arXiv preprint arXiv:2011.00597}, 2020.

\bibitem[\protect\citeauthoryear{Hinton \bgroup \em et al.\egroup
  }{2015}]{hinton2015distilling}
Geoffrey Hinton, Oriol Vinyals, and Jeff Dean.
\newblock Distilling the knowledge in a neural network.
\newblock {\em arXiv preprint arXiv:1503.02531}, 2015.

\bibitem[\protect\citeauthoryear{Jiang \bgroup \em et al.\egroup
  }{2014}]{THUMOS14}
Y.-G. Jiang, J.~Liu, A.~Roshan~Zamir, G.~Toderici, I.~Laptev, M.~Shah, and
  R.~Sukthankar.
\newblock {THUMOS} challenge: Action recognition with a large number of
  classes.
\newblock \url{http://crcv.ucf.edu/THUMOS14/}, 2014.

\bibitem[\protect\citeauthoryear{Krishna \bgroup \em et al.\egroup
  }{2017}]{krishna2017dense}
Ranjay Krishna, Kenji Hata, Frederic Ren, Li~Fei-Fei, and Juan Carlos~Niebles.
\newblock Dense-captioning events in videos.
\newblock In {\em Proceedings of the IEEE international conference on computer
  vision}, pages 706--715, 2017.

\bibitem[\protect\citeauthoryear{Lea \bgroup \em et al.\egroup
  }{2017}]{lea2017temporal}
Colin Lea, Michael~D Flynn, Rene Vidal, Austin Reiter, and Gregory~D Hager.
\newblock Temporal convolutional networks for action segmentation and
  detection.
\newblock In {\em proceedings of the IEEE Conference on Computer Vision and
  Pattern Recognition}, pages 156--165, 2017.

\bibitem[\protect\citeauthoryear{Lin \bgroup \em et al.\egroup
  }{2014}]{lin2014microsoft}
Tsung-Yi Lin, Michael Maire, Serge Belongie, James Hays, Pietro Perona, Deva
  Ramanan, Piotr Doll{\'a}r, and C~Lawrence Zitnick.
\newblock Microsoft coco: Common objects in context.
\newblock In {\em European conference on computer vision}, pages 740--755.
  Springer, 2014.

\bibitem[\protect\citeauthoryear{Lin \bgroup \em et al.\egroup
  }{2019}]{lin2019bmn}
Tianwei Lin, Xiao Liu, Xin Li, Errui Ding, and Shilei Wen.
\newblock Bmn: Boundary-matching network for temporal action proposal
  generation.
\newblock In {\em Proceedings of the IEEE International Conference on Computer
  Vision}, pages 3889--3898, 2019.

\bibitem[\protect\citeauthoryear{Mun \bgroup \em et al.\egroup
  }{2019}]{mun2019streamlined}
Jonghwan Mun, Linjie Yang, Zhou Ren, Ning Xu, and Bohyung Han.
\newblock Streamlined dense video captioning.
\newblock In {\em Proceedings of the IEEE Conference on Computer Vision and
  Pattern Recognition}, pages 6588--6597, 2019.

\bibitem[\protect\citeauthoryear{Peng \bgroup \em et al.\egroup
  }{2019}]{peng2019comic}
Xi~Peng, Zhenyu Huang, Jiancheng Lv, Hongyuan Zhu, and Joey~Tianyi Zhou.
\newblock Comic: Multi-view clustering without parameter selection.
\newblock In {\em International Conference on Machine Learning}, pages
  5092--5101, 2019.

\bibitem[\protect\citeauthoryear{Shen \bgroup \em et al.\egroup
  }{2017}]{shen2017weakly}
Zhiqiang Shen, Jianguo Li, Zhou Su, Minjun Li, Yurong Chen, Yu-Gang Jiang, and
  Xiangyang Xue.
\newblock Weakly supervised dense video captioning.
\newblock In {\em Proceedings of the IEEE Conference on Computer Vision and
  Pattern Recognition}, pages 1916--1924, 2017.

\bibitem[\protect\citeauthoryear{Song and Soleymani}{2019}]{song2019polysemous}
Yale Song and Mohammad Soleymani.
\newblock Polysemous visual-semantic embedding for cross-modal retrieval.
\newblock In {\em Proceedings of the IEEE Conference on Computer Vision and
  Pattern Recognition}, pages 1979--1988, 2019.

\bibitem[\protect\citeauthoryear{Tran \bgroup \em et al.\egroup
  }{2015}]{tran2015learning}
Du~Tran, Lubomir Bourdev, Rob Fergus, Lorenzo Torresani, and Manohar Paluri.
\newblock Learning spatiotemporal features with 3d convolutional networks.
\newblock In {\em Proceedings of the IEEE international conference on computer
  vision}, pages 4489--4497, 2015.

\bibitem[\protect\citeauthoryear{Xiong \bgroup \em et al.\egroup
  }{2018}]{xiong2018move}
Yilei Xiong, Bo~Dai, and Dahua Lin.
\newblock Move forward and tell: A progressive generator of video descriptions.
\newblock In {\em Proceedings of the European Conference on Computer Vision
  (ECCV)}, pages 468--483, 2018.

\bibitem[\protect\citeauthoryear{Yu \bgroup \em et al.\egroup
  }{2020}]{yu2020long}
Ting Yu, Jun Yu, Zhou Yu, Qingming Huang, and Qi~Tian.
\newblock Long-term video question answering via multimodal hierarchical memory
  attentive networks.
\newblock {\em IEEE Transactions on Circuits and Systems for Video Technology},
  2020.

\bibitem[\protect\citeauthoryear{Zhao \bgroup \em et al.\egroup
  }{2017}]{zhao2017temporal}
Yue Zhao, Yuanjun Xiong, Limin Wang, Zhirong Wu, Xiaoou Tang, and Dahua Lin.
\newblock Temporal action detection with structured segment networks.
\newblock In {\em Proceedings of the IEEE International Conference on Computer
  Vision}, pages 2914--2923, 2017.

\bibitem[\protect\citeauthoryear{Zhou \bgroup \em et al.\egroup
  }{2018}]{zhou2018end}
Luowei Zhou, Yingbo Zhou, Jason~J Corso, Richard Socher, and Caiming Xiong.
\newblock End-to-end dense video captioning with masked transformer.
\newblock In {\em Proceedings of the IEEE Conference on Computer Vision and
  Pattern Recognition}, pages 8739--8748, 2018.

\end{thebibliography}

\end{document}